\pdfoutput=1
\documentclass[11pt]{article}

\usepackage[]{acl}

\usepackage{times}
\usepackage{latexsym}

\usepackage[T1]{fontenc}
\usepackage[utf8]{inputenc}

\usepackage{microtype}
\usepackage{inconsolata}

\usepackage{lipsum}
\usepackage{enumitem}
\usepackage{linguex}
\usepackage{bbm}
\usepackage{booktabs}
\usepackage{hyperref}
\usepackage[dvipsnames]{xcolor}

\def\g.{\b.}
\def\h.{\b.}

\usepackage{amsmath}

\usepackage{soul}
\usepackage{xcolor}
\usepackage{graphicx}
\usepackage{cleveref}
\usepackage{utils}

\newcommand{\coord}{\textsc{coord}}
\newcommand{\arc}{\textsc{arc}}

\title{Hey, wait a minute: on at-issue sensitivity in Language Models}

\author{Sanghee J. Kim \\
  The University of Chicago \\
  \texttt{sangheekim@uchicago.edu} \\\And
  Kanishka Misra \\
  The University of Texas at Austin \\
  \texttt{kmisra@utexas.edu} \\}

\begin{document}
\maketitle
\begin{abstract}

Evaluating the naturalness of dialogue in language models (LMs) is not trivial: notions of \textit{naturalness} vary, and scalable quantitative metrics remain limited. This study leverages the linguistic notion of \textit{at-issueness} to assess dialogue naturalness and introduces a new method: Divide, Generate, Recombine, and Compare (DGRC). DGRC (i) divides a dialogue as a prompt, (ii) generates continuations for subparts using LMs, (iii) recombines the dialogue and continuations, and (iv) compares the likelihoods of the recombined sequences. This approach mitigates bias in linguistic analyses of LMs and enables systematic testing of discourse-sensitive behavior. Applying DGRC, we find that LMs prefer to continue dialogue on at-issue content, with this effect enhanced in instruct-tuned models. They also reduce their at-issue preference when relevant cues (e.g., ``Hey, wait a minute'') are present. Although instruct-tuning does not further amplify this modulation, the pattern reflects a hallmark of successful dialogue dynamics.
\end{abstract}

\section{Introduction}

Language models (LMs) have made substantial progress in dialogue quality. Findings and surveys report meaningful improvements, including in areas such as user engagement \citep{ferron2023meep} and personalization \citep{zhang2025personalization}. Nevertheless, evaluating LMs' dialogue responses remains a central challenge \citep[][for a review]{guan2025evaluating}. Dialogue varies in topic, style, and length, and human judgments are often inconsistent across evaluators. To address this, recent work used predefined evaluation criteria \citep[e.g.,][]{lin2023llm}, but even notions like \textit{naturalness} remain vague. For instance, ``Could the utterance have been produced by a native speaker?'' is a criterion used for evaluating naturalness \citep[e.g.,][]{reddy2022automating}, but it offers little guidance. To address this limitation, we propose a systematic and linguistically grounded approach to evaluating dialogue \textit{naturalness}.

Our approach is defined by two core characteristics. First, we ground \textit{naturalness} in the linguistic notion of \textit{at-issueness} \citep{potts2005, koev2022, simons2010, tonhauser2012}, which distinguishes between content that advances the discourse (\textit{at-issue}) and content that supplements it without shifting the conversational trajectory \citep{hunter2016, jasinskaja2016, riester2019}. While there have been approaches to apply the broad notion of at-issueness to discourse analysis \citep[e.g.,][]{ko2022discourse, kim-etal-2022-dialogue}, its use in evaluating modern LMs is limited.
Second, building on a minimal-pair, or templatic format widely used in evaluating syntactic, semantic, and pragmatic abilities of language models \citep{marvin2018targeted, warstadt2020blimp, misra-etal-2023-comps, hu2022predicting}, we introduce Divide, Generate, Recombine, and Compare (\textbf{DGRC}). While the existing templatic approach provides controlled contexts for testing model sensitivity, it reduces to only evaluating a pre-defined pair of sentences, which is less ideal for capturing conversational dynamics. DGRC retains the structure of minimal contrasts but replaces full-sentence pairs with multiple possible utterance fragments sampled from the model itself, differing in the content within the previous utterance they refer to. In this way, DGRC combines the strengths of templatic, surprisal-based evaluation with the flexibility of open-ended continuation while enabling a linguistic assessment of discourse dynamics, which has been less deeply considered in previous work.

We find LMs show a preference for \textit{at-issue} content: they are more likely to respond to an utterance's at-issue than its not-at-issue content. Moreover, instruct-tuned models exhibit this tendency more strongly than non-instruct-tuned models, even when controlling for a recency effect. Yet, this advantage disappears when digression cues reduce at-issue preference. These results highlight the strength of DGRC's flexible, fine-grained method for evaluating LM dialogue dynamics.

\section{Dialogue naturalness and \textit{at-issueness}}
\label{sec:at-issueness}

We treat dialogue as a single-turn pair ($U$, $R$), with $U$ as the initiating utterance and $R$ the response.
To evaluate whether a response $R$ constitutes a \textit{natural} continuation of $U$, we rely on the linguistic notion of \textit{at-issueness}.  
\textbf{At-issueness} captures if content conveys the main point of an utterance. If so, it is \textit{at-issue}; if parenthetical and backgrounded, it is \textit{not at-issue} \citep[][a.o.]{potts2005}. In \cref{ex:sally}, what is \textcolor{teal}{at-issue} in $U$ is Sally's encounter with the governor, while her dating status is \textcolor{brown}{not-at-issue}. At-issueness underlies conversational naturalness: a natural response typically targets at-issue content. Responses addressing not-at-issue content feel unnatural (e.g., ``Yes, Sally and Dave are dating''), but an explicit digression signal (e.g., ``Hey, wait a minute’’) can make them natural \citep{syrett2015}.

\ex. \label{ex:sally}
$U$: \textcolor{teal}{Sally}---\textcolor{brown}{Dave's girlfriend}---\textcolor{teal}{met the Illinois governor at a restaurant}. \\
$R$: She met the governor at a dog park. \newline \hfill (targets at-issue) \\
$R$: Hey, wait a minute, Sally's dating? \\ \hfill (targets not-at-issue with a digression signal)

Following \citet{kim-etal-2022-dialogue}, we use appositive relative clauses (\textbf{ARC}s)---embedded clauses marked by commas---as a stand-in for not-at-issue content:

\ex. \label{ex:librarian} \textcolor{teal}{The librarian}, \textcolor{brown}{[who likes pasta]$_{\text{VerbPhrase1}}$}, \textcolor{teal}{[is famous]$_{\text{VerbPhrase2}}$}!

In constructions as \cref{ex:librarian}, the \textcolor{teal}{main clause (MC)} conveys \textcolor{teal}{at-issue} content, while the embedded \textcolor{brown}{ARC} contributes \textcolor{brown}{not-at-issue} information.
ARCs are useful because: 1) they are \textit{parentheticals}, conventionally treated as not-at-issue \citep{anderbois2010,potts2005}, and 2) they embed both (not-)at-issue content within a single sentence \citep{jasinskaja2016}, avoiding the need for multi-sentence setup. We denote \textcolor{brown}{ARC} as \textcolor{brown}{VP1} and \textcolor{teal}{MC} as \textcolor{teal}{VP2} hereafter since this construction will be contrasted with a minimally different one in our experiments.

Building on background on response dynamics involving ARC structures, we distill two guiding intuitions.~First, responses are more likely to target at-issue content than not-at-issue content:
\begin{align*}
    p_{\textsf{LM}}(\textcolor{teal}{R_{\text{at-issue}}} \mid U) > p_{\textsf{LM}}(\textcolor{brown}{R_{\text{not-at-issue}}} \mid U)
\end{align*}

\noindent
Second, responses addressing at-issue content become less likely when they start with a digression signal (e.g., ``Hey, wait a minute''), relative to when this digression signal is absent:
\begin{align*}
    p_{\textsf{LM}}(\textcolor{teal}{R_{\text{at-issue}}} \mid U) > p_{\textsf{LM}}(\textcolor{teal}{R_{\text{at-issue}}} \mid U,R_{d}),
\end{align*}
where $R_d$ is a minimally added response ``header'' that signals digression, and $p_{\textsf{LM}}$ is the language model's probability. These two aspects serve as foundations for Experiments 1 and 2, respectively.

\section{Method}
\label{sec:dgrc}

\begin{figure*}[t]
    \centering
    \includegraphics[width=0.7\textwidth]{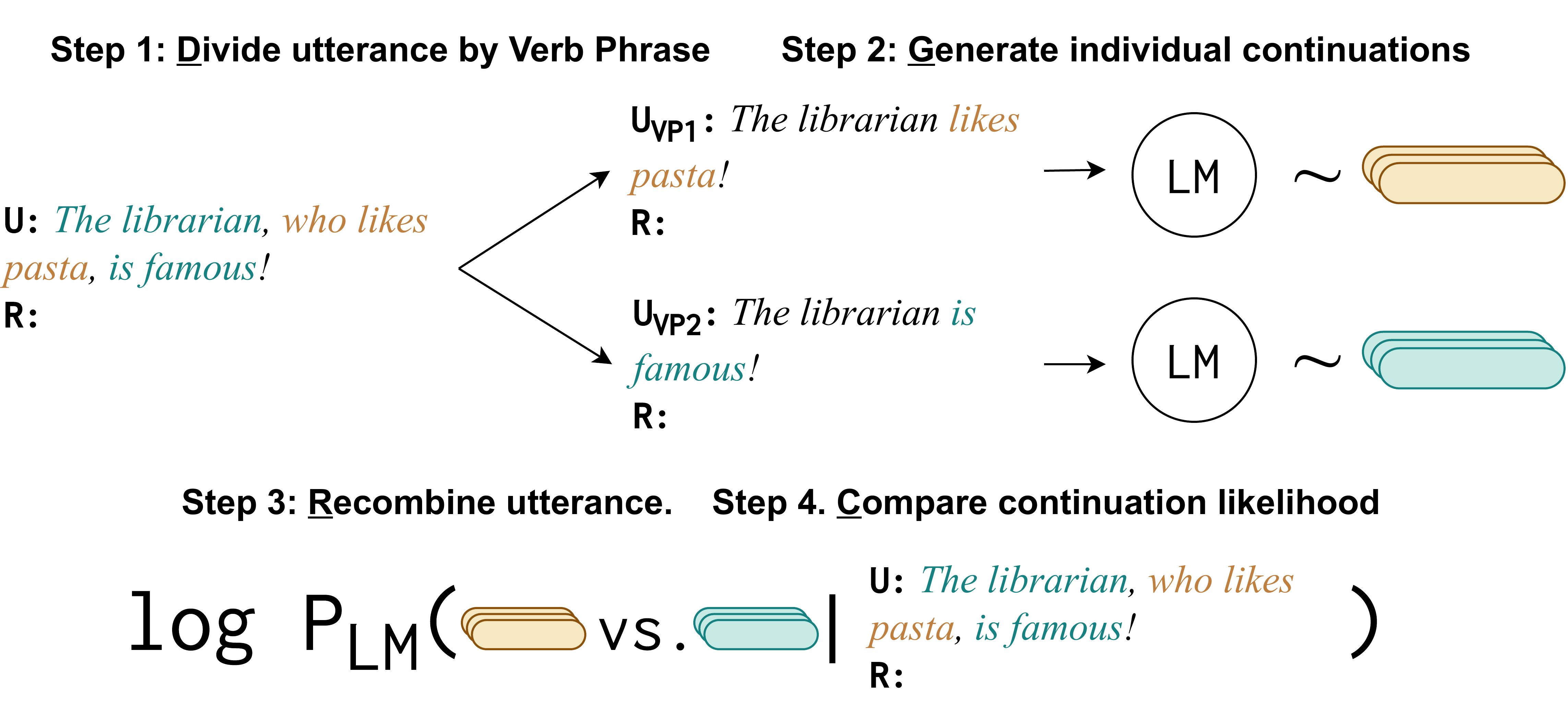}
    \caption{Visualization of the DGRC method, involving four steps: 1) Dividing the original utterance into sub-utterances; 2) Generating continuations for individual sub-utterances; 3) Recombining the sub-utterances into the original utterance, and 4) Comparing likelihoods for generated continuations. The end-result of this process allows us to characterize an LM's dialogue response dynamics. \textbf{Note:} We use log-probabilities \textit{per token} in our analyses.}
    \label{fig:dgrc}
    \vspace{-1em}
\end{figure*}

It is tempting to analyze at-issue sensitivity through minimal-pair judgments, akin to a range of linguistically motivated tests \citep{marvin2018targeted, warstadt2020blimp, misra-etal-2023-comps}. One can easily define a pair of fixed responses, each targeting a different part of the utterance, and use LM probabilities for forced-choice judgments \citep[\textit{\`a la}][]{kim-etal-2022-dialogue}. While the idea of using LM-probabilities is directly applicable, using a forced-choice paradigm is fundamentally limiting when it comes to response dynamics. First, the researcher has to select a fixed surface form as the ``correct'' response even though there could be a range of different responses that target an utterance's at-issue content. Second, even if one could hand-code a range of responses, these will be susceptible to researcher bias and may deviate from the LM's space of possible responses \citep{schuster2022sentence}.

To avoid both these limitations, we propose \textbf{Divide, Generate, Recombine, and Compare (DGRC)}, which operationalizes the LM probability comparison method as follows (also see \Cref{fig:dgrc}):
\begin{enumerate}\setlength{\itemsep}{1pt}\setlength{\parskip}{0pt}\setlength{\parsep}{0pt}
    \item We first \textbf{divide} a given $U$ (\texttt{``S VP1 VP2''}) into two independent utterances: $U_{\text{VP1}} = \texttt{``S VP1''}, U_{\text{VP2}} = \texttt{``S VP2''}$.
    \item We then prompt the LM to \textbf{generate} $n$ responses to each independent utterance: $R_{\text{VP}} = \{r^{\text{VP}}_1, \dots, r^{\text{VP}}_n\}, \text{VP} \in \{1,2\}.$
    \item We then \textbf{recombine} the independent utterances into the original one.
    \item Finally, we \textbf{compare} the (log) probabilities per token of all possible pairs of individual responses to quantify the LM's preference.
    \begin{align*}
        &\frac{1}{n^2}\sum_i^n \sum_j^n\mathbbm{1}[s(r^{\text{VP2}}_{i} \mid U) > s(r^{\text{VP1}}_{j}\mid U)],\\
        & s(x \mid U) = \log p_{\textsf{LM}}(x \mid U)/|x|
\end{align*}
\end{enumerate}

\noindent
For ARC constructions, we expect a model that is sensitive to at-issue distinctions to have a greater preference for VP2 (MC) over VP1 (ARC). By considering multiple different responses sampled \textit{from} the LM, as opposed to researcher-defined, DGRC addresses both abovementioned limitations of forced-choice tasks.

\paragraph{Data, models, and implementation}

Stimuli from \citet{kim-etal-2022-dialogue} were used as a source of utterances. The dataset contained 300 English items, each consisting of a subject NP and two VPs forming an \arc{} construction (example in \cref{ex:librarian}). For LMs: we analyzed the instruct-tuned and base versions of Llama-3-8B \citep{llama3} and Qwen2.5 families \citep{qwen2.5}, using four model sizes from 500M to 7B parameters, totaling 10 LMs. Responses were sampled using greedy decoding, and temperature, top-p, and top-k sampling. We considered several hyperparameters when applicable. We retained top-10 generations (per VP) for each utterance across all hyperparameter combinations (see DGRC Step 2), selecting those with the highest log-probabilities under the target LMs. This yielded 100 pairwise comparisons per item. For instruct-tuned models, stimuli were embedded in a chat-template format; for base models, they were embedded in a two-speaker conversational style. See \nameref{sec:appendix} for stimuli, hyperparameters, and prompts. Code and data are available at: \url{https://github.com/sangheek16/hey-wait-a-minute}.

\section{Experiments}

\subsection{Experiment 1: At-issueness preference}

In our first experiment, we analyze the extent to which LMs demonstrate sensitivity to at-issue content. We do so by performing DGRC on our dataset of \arc{} sentences and measuring at-issue preferences as described in \Cref{sec:dgrc}. To place these results in context, we follow \citet{kim-etal-2022-dialogue}, and compare these preferences to those obtained by performing DGRC on coordination structures (\coord), formed by conjoining the two VPs in our stimuli:

\ex. \label{ex:coord} The librarian [likes pasta]$_\text{VP1}$ and [is famous]$_\text{VP2}$.

This comparison is important since \coord{} structures differ from \arc{} minimally in terms of surface form (cf. \cref{ex:librarian}), and importantly include \textit{both} VPs as part of the ``main point'' of the utterance---i.e., they lose the at-issue distinction. This allows us to control for recency bias: if LMs are simply showing sensitivity to the VP2 because it is more recent, then we should observe similar behavior in both \arc{} and \coord{}. On the other hand, if LMs are non-trivially sensitive to at-issueness, then their VP2 preference should be greater in \arc{} than in \coord{}. Outside of this comparison, we also include cases where we swapped the two VPs in our stimuli, to control for potential robustness issues.

\paragraph{Results} 

\begin{figure}[t]
    \centering
    \includegraphics[width=\columnwidth]{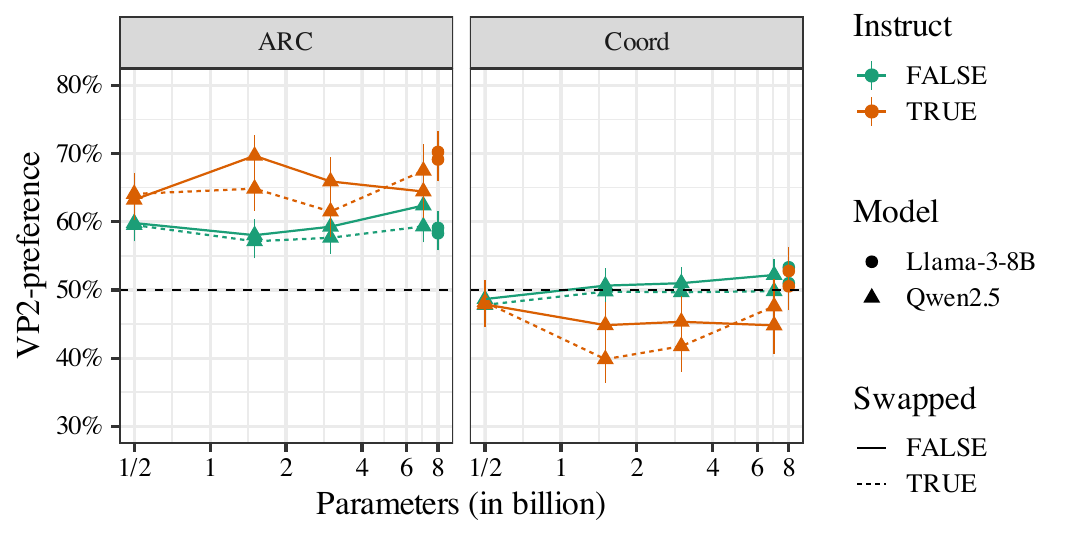}
    \caption{Experiment~1 results. VP2-preference of LMs (organized by parameter count) across \textsc{arc} and \textsc{coord} constructions, training mode (base/instruct), model family, and whether VP order was swapped.
    }
    \label{fig:exp1}
\end{figure}

\Cref{fig:exp1} shows our results.
We found LMs' preference for VP2 to be greater in \arc{} structures than in \coord{} structures 
(see \nameref{sec:appendix} for full results).
Additionally, instruct-tuned versions of an LM were consistently better in their at-issue preference than base models, suggesting a potential \textit{pragmatically sensitive} behavior that seems to arise from instruction tuning. These findings were found to be statistically significant ($p < .01$ for both), shown using a linear mixed-effects model (LMEM) analysis. No notable effect of scale was found. Finally, LMs were generally robust to VP order, as shown by the lack of a swapping effect in the LMEM analysis ($p = 0.42$). We conclude that \textbf{LMs, especially instruct-tuned ones, show sensitivity to at-issue content in dialogues}.

\subsection{Experiment 2: Sensitivity to digression}

\begin{figure}[t]
    \centering
    \includegraphics[width=\columnwidth]{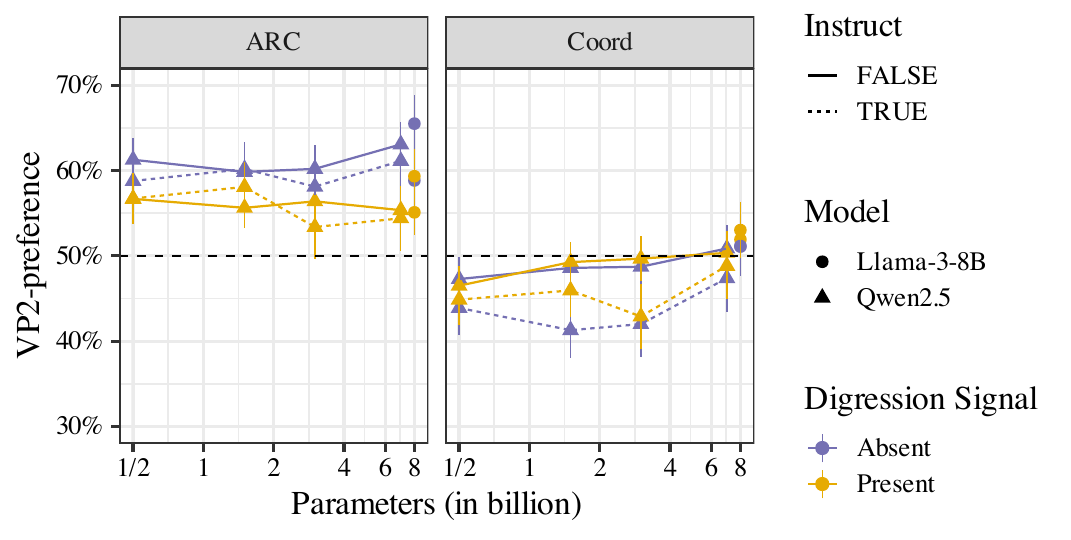}
    \caption{Experiment~2 results. VP2-preference of LMs (organized by parameter count) across \textsc{arc} and \textsc{coord} constructions, training mode (base/instruct), model family, and digression signal---absent (``No, that's not true''); present (``Hey, wait a minute'').
    }
    \label{fig:exp2}
\end{figure}

How do LMs' sensitivity to at-issue content vary in the presence of subtle cues shown to affect human at-issue sensitivities \citep{syrett2015}? For this, we turn to analyzing if LMs' at-issue sensitivity in \arc{} structures \textit{decreases} in the presense of digression, as established in \cref{sec:at-issueness}. To test this, we use digression-signaling cues such as ``Hey, wait a minute'' as response headers, following widely-used linguistic diagnoses \citep[\textit{peripherality test};][]{koev2022, amaral2007, shanon1976}. We compare LM preferences for at-issue content in responses beginning with ``Hey, wait a minute'' versus ``No, that’s not true!'', the latter controlling for general \textit{rejection}. That is, when rejecting part of the prior utterance, a direct rejection (``No'') primarily targets at-issue content \citep{murray2014, amaral2007, anderbois2010}, whereas ``Hey, wait a minute'' is expected to attenuate this, signaling rejection of the not-at-issue content \citep{syrett2015}. Since both responses reject prior content, we include ``No, that's not true!'' header in Step 2 of DGRC in order for models to contradict the content of individual VPs. As before, we test both \arc{} and \coord{} structures.

\paragraph{Results} 
\Cref{fig:exp2} shows our results. 
We found an interaction between digression signal (absent vs.~present) and structure (\arc{} vs.~\coord{}), with $p < .001$ (see \Cref{fig:exp2-interaction} for interaction plot and full results). 
In particular, LMs strongly preferred at-issue content (VP2) in the \textit{absence} of digression than in its presence, but \textit{only} in \arc{} structures and not in \coord{}, in line with our expectations.
At the same time, they also showed an overall bias towards for VP2 content in the \arc{} structure (relative to \coord{}). This suggests that digression does not always \textit{shift} preference to not-at-issue content, as it does in humans \citep[e.g.,][]{syrett2015}. In short, \textbf{LMs prefer at-issueness, with the digression-signalling header modulating the effect only when a (not-)at-issue divide is present}. We did not find a particular benefit of instruct-tuning in \arc{} structures ($p = .39$), as both types of models showed a similar pattern of VP2-preference, i.e., response targeting at-issue content.

\section{Conclusion}

We introduced DGRC, a linguistically motivated method for evaluating dialogue naturalness through the lens of at-issueness. Our results showed that LMs consistently prefer at-issue over not-at-issue content, with this tendency especially pronounced in instruct-tuned models (Experiment~1), suggesting that instruct-tuning can induce pragmatic sensitivity even without targeted training. Experiment~2 replicated models' the at-issue preference, while showing that this tendency is evident only in the context where (not-)at-issue divide is present, and that preference can be modulated when signaled with a digression header. These findings imply that enhancing sensitivity to at-issueness and dialogue dynamics could substantially improve the naturalness of model-generated dialogue. Importantly, DGRC enabled systematic, quantifiable evaluation without reliance on human evaluators or researcher-crafted prompts, complementing existing evaluation methods and allowing more fine-grained assessments of LMs in dialogues.

\vspace{3.5em}

\section*{Limitations}

\paragraph{Scope of construction} The current paper examined a specialized construction, namely a structure involving an embedded ARC. We acknowledge that this is a highly specific linguistic form, and our scope is therefore limited. However, as noted in \Cref{sec:at-issueness}, this construction offers a systematic and controlled way to compare at-issue and not-at-issue content without requiring multi-turn utterances, even when testing the intricate distinction. Moreover, findings based on this construction highlight the importance of at-issueness in evaluating the naturalness of dialogue. Yet, the single-turn dialogue setup was primarily chosen to simplify stimulus design and to allow tighter experimental control (e.g., avoiding potential confounds due to variation in construction). In future work, we plan to explore this phenomenon more deeply in full, multi-turn dialogues using the DGRC method.

\paragraph{At-issueness of constructions}
We used coordination constructions (\texttt{VP1 and VP2}) as a baseline, assuming that they do not encode the (not-)at-issue distinction within a sentence. We note that subtle differences may still arise in the extent to which each VP contributes to discourse progression, given evidence that sentence-final or more recent discourse entities have greater potential to continue the discourse \citep[e.g.,][]{frazier2005, anderbois2015, jasinskaja2016, hunter2016}. Additionally, while the ARC construction was used to create short dialogue contexts embedding an at-issue vs.\ not-at-issue distinction, linguistic theories suggest that this distinction can weaken depending on the ARC's sentential position \citep{anderbois2015}. A sentence-final ARC may behave more ``at-issue-like'' as it stands closer to the following discourse content. We did not test this configuration, and exploring this remains as future work, which will offer an even finer-grained evaluation method on dialogue dynamics.

\paragraph{Understanding instruct-tuning} Instruct-tuned models are often considered better at capturing conversational goals \citep[e.g.,][]{zhang2023instruction}. We also found such patterns in Experiment~1, where instruct-tuned models aligned more with the expected behavior. But this was not the case for Experiment~2. Recent studies suggest that instruction tuning does not necessarily yield stronger model–human alignment, when evaluated against human judgments and behavioral data \citep{zhang2023instruction, kauf2024comparing, aw2024instructiontuning, kim2025discourse}. Given these mixed findings, a principled analysis of how instruct-tuning affects human–machine dialogue alignment would be especially informative for future direction.

\section*{Acknowledgments}

ChatGPT was used for paraphrasing and spell-checking at the proofreading stage.

\bibliography{anthology, custom, all_bibliography}

\bibliographystyle{acl_natbib}

\appendix

\section{Appendix}\label{sec:appendix}

\subsection*{Data and implementation}

\paragraph{Stimuli} A sample set of data used for DGRC is shown in Table~\ref{tab:data}. These were taken from \citet{kim-etal-2022-dialogue}, which was released using the MIT License, as specified on their github.\footnote{\url{https://github.com/sangheek16/dialogue-response-dynamics}}

\begin{table*}[t]
\centering
\begin{tabular}
{p{0.18\linewidth}p{0.37\linewidth}p{0.37\linewidth}}
\toprule
Subject NP & VP1 & VP2 \\
\midrule
The nurse & met the Illinois governor at a Greek restaurant & looks confident \\
The athlete & went to the post office & has been checking the clock for five hours \\
The programmer & used to drink three cups of coffee every day & is about to fall asleep \\
The anthropologist & went out for a date & was solving a crossword puzzle \\
The nanny & seemed very eager to return home & would wear glasses in the day time \\
The overseer & reads eight books a month & hired a teenager to mow the lawn \\
The nun & rides a bike to the nearest park & has been in town for 10 years \\
The senator & finds humor in the worst situations & is not good at riding a bike \\
The waitress & wears a fancy watch & was upset about the cleanliness of City Hall \\
The make-up artist & occasionally writes a blog about historical figures & would drive to the nearest library when the weather is bad \\
The model & has been dreaming of buying a private jet & went to the post office \\
The receptionist & has been earning money by making YouTube videos & has lots of friends \\
The producer & has been around for a while & is standing next to the tree \\
The detective & has been chosen to audition for the All Stars game & was angry \\
The inspector & has been checking the clock for five hours & would wake up early on Christmas Eve \\
The fisherman & is good at communication & got stung by a wasp the other day \\
The diver & is reading a newly released sci-fi book & takes a vitamin every day \\
The anthropologist & is about to fall asleep & has been collecting fridge magnets for five years \\
The psychologist & is good at board games & was singing a song \\
The pilot & is subscribed to seventy different newsletters & would read books at a park nearby \\
The tenant & was trying to fix the broken window & seemed very eager to return home \\
The farmer & was disappointed with the weather & occasionally takes a nap after lunch \\
The overseer & was happy about the cross-country road trip & likes bungee jumping \\
The linguist & was disappointed with the weather & has been staying in Hawaii for two months \\
The bartender & was mentioned in the newspaper & is extremely fickle and demanding \\
The hairdresser & was unhappy about all the noise on the streets & would make pasta for dinner \\
The painter & would drive to the nearest library when the weather is bad & suggested a good cleaning company to Ashley \\
The musician & would wear a yellow hat on sunny days & enjoys hiking \\
The lecturer & would have salad and boiled eggs for lunch & has been a fan of Rihanna since her debut \\
The administrator & would go to the movies every week & is never late \\
The writer & would swim in the Lake on Monday mornings & was being chased by a few people\\
\bottomrule
\end{tabular}
\caption{Examples of subject NPs with VP1 and VP2 used for generating experimental stimuli.}
\label{tab:data}
\end{table*}

\paragraph{Models and hyperparameters}

We evaluated instruct-tuned and base models from the Llama-3-8B and Qwen2.5 family:

\begin{itemize}\setlength{\itemsep}{1pt}\setlength{\parskip}{0pt}\setlength{\parsep}{0pt}
	\item Instruct-based models: Meta-Llama-3-8B-Instruct, Qwen2.5-0.5B-Instruct, Qwen2.5-1.5B-Instruct, Qwen2.5-3B-Instruct, and Qwen2.5-7B-Instruct
	\item Base models: Meta-Llama-3-8B, Qwen2.5-0.5B, Qwen2.5-1.5B, Qwen2.5-3B, and Qwen2.5-7B
\end{itemize}

For each model, generations were produced using combinations of the following hyperparameters: top-$p$ where $p \in \{0, 0.9, 0.95\}$, top-$k$ values $k \in \{50, 0\}$, and temperatures $t \in \{0.7, 1.0\}$. The top-10 generations out of all generations were selected based on the log probabilities of the utterance--response sequence. All log probabilities are computed using minicons \citep{misra2022minicons}. Models were run on a compute cluster containing a single NVIDIA RTX6000 Ada GPU, with 48GB RAM. 

\paragraph{Implementation}
For instruct-tuned models, we used the model's chat template, which takes a sequence of roles: \texttt{system}, \texttt{user}, and, for Experiment~2, additionally \texttt{assistant}. The \texttt{system} message provided a short instruction, as shown below. The divided utterance was placed in the \texttt{user} content. In Experiment~2, headers were inserted as content of \texttt{assistant} (e.g., ``No, that's not true!'' or ``Hey, wait a minute!''). The following is an example:

\begin{quote}
			[\texttt{system}] ``Please respond to the following message as naturally as possible, using a single sentence,  as if we were talking to each other.'' Please keep it short. \newline 
			[\texttt{user}] ``The librarian likes pasta.'' \newline
			[\texttt{assistant}] ``No, that's not true!''
\end{quote}

For base models, which do not support chat formatting, we instead included the utterance and response as a two-person dialogue using direct quotation, similar to \citet{kim-etal-2022-dialogue}: \texttt{\$name1} said, ``\texttt{\$stimulus},'' and \texttt{\$name2} replied, (``\texttt{\$header}''). The header (in parentheses) was included only in Experiment~2. Placeholders for \texttt{\$name1} and \texttt{\$name2} were replaced with proper names (e.g., Marco, Ellie), randomly sampled from a list of 400 names. An example is shown below: 

\begin{quote}
    Marco said, ``The librarian, who likes pasta, is famous,'' and Ellie replied, (``No, that's not true!'')
\end{quote}

\subsection*{Linear Mixed-Effects Modeling Results}

We analyzed results from our experiments using linear mixed-effects models using the \texttt{lme4} \citep{bates2015fitting} and \texttt{lmerTest} \citep{kuznetsova2017lmertest} packages in R. In what follows we describe the model formula and show interaction plots (when applicable).

\paragraph{Experiment~1} We used the following formula to analyze our results in this experiment:

\small
\begin{align*}
    &\texttt{vp2\_pref} \sim \texttt{swapped} + \texttt{instruct} \times \texttt{structure}\\& +(1 + \texttt{swapped} + \texttt{instruct} \times \texttt{structure} \mid \texttt{item}) \\&+(1 \mid \texttt{model}),
\end{align*}
\normalsize
where \texttt{swapped} indicates whether or not the VPs were swapped, \texttt{instruct} indicates the presence/absence of instruct-tuning, \texttt{structure} indicates the type of structure (\arc{} vs. \coord{}), \texttt{model} is the model, and \texttt{item} is the individual item (which determines what lexical items occur in the structure---i.e., the lexical content of NP, VP1, VP2). Our results are shown in \Cref{tab:exp1-lmer}.

\begin{table*}[t]
\centering
\begin{tabular}{@{}rrrrrr@{}}
\toprule
\textbf{Term} & \textbf{Estimate} & \textbf{Std. Error} & \textbf{df} & \textbf{$t$} & \textbf{$p$} \\ \midrule
(Intercept) & 0.56 & 0.01 & 16.08 & 54.06 & $< \text{.001}$ \\
swapped & -0.01 & 0.01 & 298.92 & -0.73 & 0.46 \\
instruct & 0.01 & 0.00 & 11684.73 & 3.45 & $< \text{.001}$ \\
mode & 0.14 & 0.00 & 9214.68 & 32.84 & $< \text{.001}$ \\
instruct:mode & 0.11 & 0.01 & 11117.55 & 12.78 & $< \text{.001}$ \\ \bottomrule
\end{tabular}%
\caption{Linear Mixed Effects Model summary for Experiment 1.}
\label{tab:exp1-lmer}
\end{table*}

\begin{figure}[t]
    \centering
    \includegraphics[width=0.8\columnwidth]{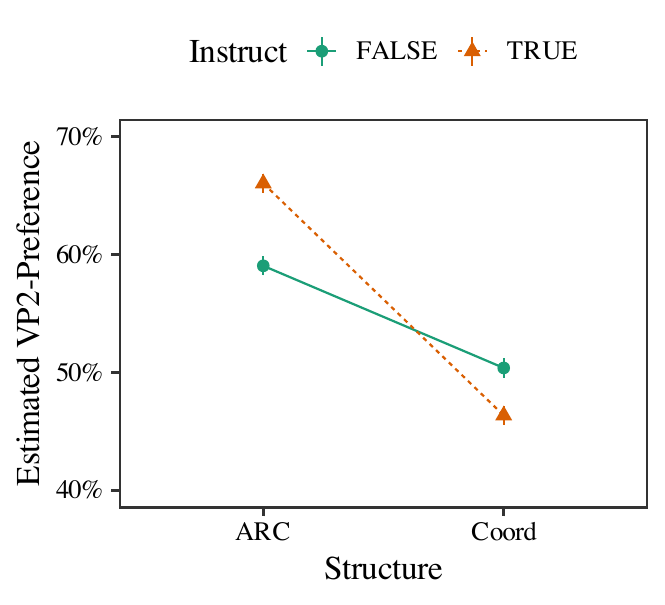}
    \caption{Effect of instruct-tuning on results in Experiment~1. VP2-preference of LMs across \arc{} and \coord{} structures, modulated by instruct-tuned vs. base model.}
    \label{fig:exp1-interaction}
    \vspace{-1em}
\end{figure}

The interaction between instruct-tuning and structure---found to be significant---is shown in \Cref{fig:exp1-interaction}. The VP2-preference, i.e., targeting at-issue content, is salient in \arc{} constructions, particularly with instruct-tuned models.

\paragraph{Experiment~2}

We used the following formula to analyze our results in this experiment:

\small
\begin{align*}
    &\texttt{vp2\_pref} \sim \texttt{header} \times \texttt{instruct} \times \texttt{structure}\\& +(1 + \texttt{header} \times \texttt{instruct} \times \texttt{structure} \mid \texttt{item}) \\&+(1 + \texttt{header} \times \texttt{instruct} \times \texttt{structure} \mid \texttt{model}),
\end{align*}
\normalsize
where \texttt{header} indicates the response header of the model (1 if digression is present---i.e., ``Hey, wait a minute!'', and 0 if it is absent---i.e., ``No, that's not true!''), \texttt{instruct} indicates the presence/absence of instruct-tuning, \texttt{structure} indicates the type of structure (\arc{} vs. \coord{}), \texttt{model} is the model, and \texttt{item} is the individual item (which determines what lexical items occur in the structure---i.e., the lexical content of NP, VP1, VP2). Our results are shown in \Cref{tab:exp2-lmer}.

\begin{table*}[t]
\centering
\begin{tabular}{@{}rrrrrr@{}}
\toprule
\textbf{Term} & \textbf{Estimate} & \textbf{Std. Error} & \textbf{df} & \textbf{$t$} & \textbf{$p$} \\ \midrule
(Intercept) & 0.53 & 0.01 & 12.99 & 49.81 & $< \text{.001}$ \\
header & 0.02 & 0.01 & 5.18 & 3.14 & 0.02 \\
instruct & -0.01 & 0.02 & 13.03 & -0.88 & 0.40 \\
structure & 0.11 & 0.01 & 4.10 & 11.21 & $< \text{.001}$ \\ \midrule
header:instruct & -0.01 & 0.01 & 11385.99 & -1.38 & 0.17 \\
header:structure & 0.06 & 0.01 & 11385.99 & 7.17 & $< \text{.001}$ \\
instruct:structure & 0.04 & 0.01 & 11385.99 & 4.57 & $< \text{.001}$ \\
header:instruct:structure & 0.01 & 0.02 & 11385.99 & 0.80 & 0.42 \\ \bottomrule
\end{tabular}%
\caption{Linear Mixed Effects Model summary for Experiment 2.}
\label{tab:exp2-lmer}
\end{table*}

\begin{figure}[t]
    \centering
    \includegraphics[width=0.8\columnwidth]{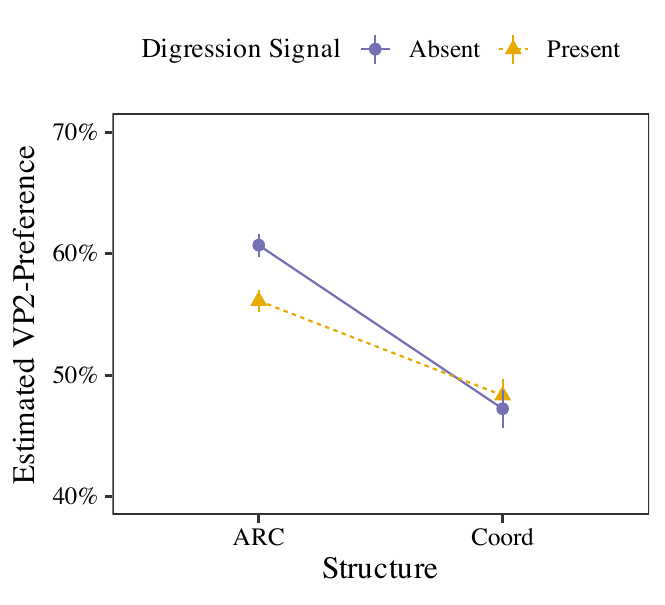}
    \caption{Effect of the digression signal in Experiment~2. VP2-preference of LMs by digression signal, modulated by \arc{} and \coord{} structures.}
    \label{fig:exp2-interaction}
\end{figure}

We found a significant interaction between digression signal and structure in Experiment~2, as shown in \Cref{fig:exp2-interaction}. A digression header signaling that the response would target not-at-issue content led to a reduced preference for VP2, i.e., at-issue content. This effect disappeared in the \coord{} structure, suggesting that the influence of a digression signal is evident only when a division between at-issue and not-at-issue content is present, namely, in the \arc{} structure.

\end{document}